\documentclass[letterpaper, 10 pt, conference]{ieeeconf}  

\IEEEoverridecommandlockouts                              
\title{Parametric Value Approximation for General-sum Differential Games with State Constraints}

\author{Lei Zhang$^{1}$, Mukesh Ghimire$^{1}$, Wenlong Zhang$^{2}$, Zhe Xu$^{1}$, Yi Ren$^{1}$
\thanks{This work was supported by the National Science Foundation under Grants CMMI-1925403 and NSF CNS-2101052.}
\thanks{$^{1}$L. Zhang, M. Ghimire, Z. Xu, and Y. Ren are with Department of Mechanical and Aerospace Engineering, Arizona State University, Tempe, AZ, 85287, USA. Email:
        {\tt\small \{lzhan300, mghimire, yiren\}@asu.edu}}
\thanks{$^{2}$W. Zhang is with School of Manufacturing Systems and Networks, Ira A. Fulton Schools of Engineering, Arizona State University, Mesa, AZ, 85212, USA. Email:
        {\tt\small wenlong.zhang@asu.edu}}%
}

\usepackage[utf8]{inputenc}
\usepackage{amssymb}
\usepackage{amsmath}
\usepackage{graphicx}  
\usepackage{xcolor}
\usepackage[colorlinks=true]{hyperref}
\usepackage[noadjust]{cite}
\DeclareMathOperator*{\argmax}{arg\,max}

\usepackage[linesnumbered,ruled]{algorithm2e} 
\usepackage{color}

\usepackage{graphicx}       
\usepackage{wrapfig}
\usepackage{subfigure}     

\usepackage{booktabs}
\usepackage{dsfont}
\usepackage{bbm}
\usepackage{caption}
\usepackage{multirow}
\usepackage[font=footnotesize]{caption}

\iftrue 

        \newcommand{\cutparagraphup}{\vspace*{-0.17in}}
        \newcommand{\cutparagraphdown}{\vspace*{-0.03in}}

        \newcommand{\cutcaptionup}{\vspace*{-0.1in}}
        \newcommand{\cutcaptiondown}{\vspace*{-0.2in}}

        \newcommand{\cutequationup}{\vspace*{-0.07in}}
        \newcommand{\cutequationdown}{\vspace*{-0.07in}}

        \newcommand{\cuttableup}{}
        \newcommand{\cuttabledown}{}

        \newcommand{\cut}{{\vspace*{-0.02in}}}
        \newcommand{\cutmore}{{\vspace*{-0.06in}}}
        \newcommand{\negcut}{}

\else 

        \newcommand{\cutparagraphup}{
        \newcommand{\cutparagraphdown}{}

        \newcommand{\cutcaptionup}{}
        \newcommand{\cutcaptiondown}{}

        \newcommand{\cutequationup}{}
        \newcommand{\cutequationdown}{}

        \newcommand{\cuttableup}{}
        \newcommand{\cuttabledown}{}

        \newcommand{\cut}{}
        \newcommand{\cutmore}{}
        \newcommand{\negcut}{}
\fi

\begin{document}
\maketitle
\thispagestyle{empty}
\pagestyle{empty}

\begin{abstract}
General-sum differential games can approximate values solved by Hamilton-Jacobi-Isaacs (HJI) equations for efficient inference when information is incomplete. However, solving such games through conventional methods encounters the curse of dimensionality (CoD). Physics-informed neural networks (PINNs) offer a scalable approach to alleviate the CoD and approximate values, but there exist convergence issues for value approximations through vanilla PINNs when state constraints lead to values with large Lipschitz constants, particularly in safety-critical applications. In addition to addressing CoD, it is necessary to learn a generalizable value across a parametric space of games, rather than training multiple ones for each specific player-type configuration. To overcome these challenges, we propose a Hybrid Neural Operator (HNO), which is an operator that can map parameter functions for games to value functions. HNO leverages informative supervised data and samples PDE-driven data across entire spatial-temporal space for model refinement. We evaluate HNO on 9D and 13D scenarios with nonlinear dynamics and state constraints, comparing it against a Supervised Neural Operator (a variant of DeepONet). Under the same computational budget and training data, HNO outperforms SNO for safety performance. This work provides a step toward scalable and generalizable value function approximation, enabling real-time inference for complex human-robot or multi-agent interactions.
\end{abstract}

\section{Introduction}
\label{sec:intro}
Many human-robot interactions (HRIs) or multi-agent interactions (MAIs) can be formulated as differential games with complete information~\cite{spica2020real,zhang2023approximating} or incomplete information~\cite{ghimire2024state,muchen2024mixed}. There are substantial efforts for solving differential games with complete information using numerical approaches (e.g., iterative LQR~\cite{fridovich2020efficient}, nested KKT~\cite{schwarting2019social}, augmented Lagrangian~\cite{cleac2019algames}). However, it remains a challenge to achieve real-time inference, particularly for incomplete-information settings, where existing solvers only operate at most 60 Hz~\cite{cleac2019algames}, while applications such as safety-critical HRIs require inference speed at around 500 Hz~\cite{zhang2024value}. A potential solution is to approximate Nash equilibrium values, 
which are viscosity solutions to Hamilton-Jacobi-Isaacs (HJI) equations~\cite{viscosity, mitchell2005time}. However, solving HJI equations through conventional approaches suffers from the curse of dimensionality (CoD)~\cite{mitchell2003}, which makes them computationally intractable for high-dimensional systems. The emergence of physics-informed neural networks (PINNs) provides a scalable approach, which uses machine learning to approximate PDE solutions and alleviates the challenges of CoD. Although using PINNs to approximate value has some promising progress~\cite{darbon2020overcoming, mukherjee2023bridging}, several key challenges still remain: First, HRIs or MAIs often involve state constraints for safety specification. Approximating values with vanilla PINNs becomes challenging because these constraints lead to large Lipschitz constants in the values~\cite{zhang2024value}. Specifically, when players at some states and time cannot avoid collisions, the corresponding values become infinite, which makes convergence difficult. 
Second, accurate value approximations require not only a low approximation error but also a precise computation of the value gradient with respect to states~\cite{yu2022gradient}. Since control policies are derived from value functions, a poor approximation of the value landscape can lead to unsafe control policies. Lastly, in incomplete-information settings, a player needs to update its belief about the player-type parameters of other players during interactions. To enable efficient inference without model retraining, approximated values and policies should generalize well across parameter space.
In this paper, we specifically examine parametric value functions with state constraints, where these constraints are also parametrized by player types and lead to large penalties.

To overcome the first two challenges, we adopt the hybrid PINN framework proposed in~\cite{zhang2024value}, which demonstrates good safety and generalization performance. The advantage of this approach is to use informative supervised data to capture the location of discontinuity for the values and their gradients. Additionally, it samples states and time across the entire state-time space to satisfy PDE residuals and boundary conditions to improve convergence. For the last challenge, we propose a Hybrid Neural Operator (HNO), inspired by physics-informed DeepONet (PINO)~\cite{wang2021learning}, to approximate parametric values and systematically investigate their safety performance across parameter space. While Pontryagin Neural Operator (PNO)~\cite{zhang2024pontryagin} demonstrates lower collision rates in 5D linear dynamics, there are no empirical studies for higher-dimensional nonlinear dynamics. In this work, we fix this gap and evaluate the efficacy of HNO for 9D and 13D case studies with nonlinear dynamics.

Our main contributions are as follows: (1) We conduct a comprehensive assessment of safety performance across three high-dimensional nonlinear dynamics for complete-information games with state constraints, comparing HNO with Supervised Neural Operator (SNO)—a variant of DeepONet~\cite{lu2021learning}; (2) Under the same computational budgets and training data points, we empirically demonstrate that HNO consistently outperforms SNO in safety performance for both seen and unseen player-type parameters; (3) We investigate the role of activation function selection in neural operator training. Our results show that the smooth and differentiable \texttt{tanh} activation achieves better and more robust safety performance compared to \texttt{sin} and \texttt{relu}.

\section{Related Work}
\label{sec:related}
\textbf{Differential games with state constraints.} The existence of values for pursuit-evasion differential games (a subclass of zero-sum games) with state constraints and complete information are derived in~\cite{cardaliaguet2000pursuit}. However, solving such problems remains challenging, as state constraints typically induce discontinuities in the value function~\cite{altarovici2013general}. To address this issue, epigraphical techniques are introduced to convert discontinuous value functions into continuous ones, facilitating value computation in zero-sum settings~\cite{lee2021hamilton, gammoudi2023differential}. Recent work~\cite{zhang2024value} extends the epigraphical approach from zero-sum games with state constraints and complete information to general-sum settings and proves the existence of viscosity solutions to HJI equations. 

\textbf{HJI equations and PINNs.}
HJI equations, a class of first-order parabolic nonlinear PDEs, are the general mathematical formulations for solving differential games. However, conventional numerical approaches (e.g., level set methods~\cite{osher2004level, mitchell2005toolbox}, essentially non-oscillatory schemes~\cite{osher1991high}) suffer from CoD, making high-dimensional problems computationally intractable. To overcome these challenges, PINNs are put forward to solve high-dimensional PDEs, including HJ equations, and leverage their Monte Carlo nature to circumvent CoD when the solution is smooth~\cite{weinan2021algorithms}. PINNs and their variants incorporate PDE knowledge into the neural network training process, optimizing the loss function based on boundary residual~\cite{han2020convergence, han2018solving}, PDE residual~\cite{jagtap2020adaptive,deepreach} and supervised data from ground truth solutions~\cite{nakamura2021}. Recent studies prove the convergence of PINNs for problems with smooth solutions~\cite{han2020convergence, shin2020convergence, ito2021neural} and derive generalization error bounds using the Neural Tangent Kernel (NTK) framework~\cite{lau2024pinnacle}. While PINNs have promising progress in approximating discontinuous solutions~\cite{krishnapriyan2021characterizing, jagtap2020adaptive, mojgani2023kolmogorov}, solving PDEs with only initial or terminal conditions remains an open challenge, particularly for HJ equations with state constraints~\cite{zhang2024value}.

\textbf{Neural operators.}
Neural operators are novel methods designed to learn mappings between function spaces~\cite{kovachki2023neural}, making them well suited to solve parametric PDEs~\cite{wang2021learning}. The first work of neural operator is DeepONet~\cite{lu2021learning}, which employs a branch-trunk architecture to extend the universal approximation theorem~\cite{chen1995universal}. In this framework, the branch network extracts key features from input functions that represent PDE parameters, and the trunk network learns the basis functions that compose parametric PDE solutions. PINO~\cite{wang2021learning} incorporates PDE and boundary residuals into DeepONet and extends neural operators from supervised learning to PINN learning. This work demonstrates the efficacy of PINO in solving parametric physics equations on 2D and 3D state spaces. For value approximation in differential games, pointwise function approximation (e.g., DeepONet and PINO) is more appropriate than learning the entire function (e.g., FNO~\cite{li2020fourier}, LNO~\cite{wang2024latent}, GNOT~\cite{hao2023gnot}, and MgNO~\cite{he2023mgno}), as approximated values are used as closed-loop controllers at specific input points.

\section{Methods}
\label{sec:diff}

\textbf{Notations and assumptions.} For a two-player general-sum differential game with complete information, we consider time-invariant dynamics. Player $i$ follows system dynamics as $\dot{x}_i = f_i(x_i, u_i)$, where $x_i \in \mathcal{X}_i \subseteq \mathbb{R}^{d_x}$ is the system state and $u_i \in \mathcal{U}_i \subseteq \mathbb{R}^{d_u}$ is the control input. We define $\textbf{a}_i=(a_i, a_{-i})$, which involves any element $a_i$ from Player $i$ and $a_{-i}$ from the fellow Player $-i$. We also denote the joint state space for both players as $\mathcal{X} = \bigcup_{i=1,2}\mathcal{X}_i$. The instantaneous loss for Player $i$ is denoted as $l_i(x_i, u_i): \mathcal{X}_i \times \mathcal{U}_i \rightarrow \mathbb{R}$ and the terminal loss is $g_i(x_i): \mathcal{X}_i \rightarrow \mathbb{R}$. The game is played over a finite time horizon $[0, T]$. Player $i$ follows a policy $\alpha_i \in \mathcal{A}$, satisfying the mapping $\mathcal{X} \times [0,T] \rightarrow \mathcal{U}_i$. Given policy $\alpha_i$, dynamics $f_i$, and any initial condition $(x_i, t)$, the state of Player $i$ at time $s$ is denoted as $x_s^{x_i,t,\alpha_i}$ and the joint state at $s$ is $\textbf{x}_s^{\textbf{x}_i,t,\boldsymbol{\alpha}_i}:=\left(x_s^{x_i,t,\alpha_i}, x_s^{x_{-i},t,\alpha_{-i}}\right)$. We introduce $c_i(\textbf{x}_i): \mathcal{X} \rightarrow \mathbb{R}$ as a state penalty derived from Player $i$'s state constraints: $c_i = 0$ if $\textbf{x}_i$ satisfies Player $i$'s state constraints, or otherwise $c_i$ is a large positive number. In this study, $c_i$ is differentiable but has a large Lipschitz constant. The value function of Player $i$ is denoted as $\vartheta_i(\textbf{x}_i, t): \mathcal{X} \times [0,T] \rightarrow \mathbb{R}$. Furthermore, we introduce player-type parameter $\theta \in \Theta \subseteq \mathbb{R}^{d_{\theta}}$, where $\Theta$ represents the type space. Consequently, $f_i$, $l_i$, $g_i$, $c_i$, and $\vartheta_i$ become $\theta$-dependent functions (e.g., $l_i^{\theta}$ denotes the instantaneous loss function of Player $i$ with type $\theta$). We make the assumptions as follows: $\mathcal{U}_i$ is compact and convex; $f_i$ and $c_i$ are Lipschitz continuous; $l_i$ and $g_i$ are Lipschitz continuous and bounded.   

\textbf{Value and HJI with state constraints}. We define the payoff function for Player $i$ under a given policy pair $\boldsymbol{\alpha}$
\begin{equation}
\begin{aligned}
 J_i(\textbf{x}_i, t, \boldsymbol{\alpha}_i) := & \int_t^T \big(l_i\left(x_s^{x_i,t,\alpha_i}, \alpha_i\left(\textbf{x}_s^{\textbf{x}_i,t,\alpha_i,\alpha_{-i}}, s\right)\right) \\ 
 & + c_i(\textbf{x}_s^{\textbf{x}_i,t,\alpha_i,\alpha_{-i}}) \big)ds + g_i\left(x_T^{x_i,t,\alpha_i}\right)
\label{eq:value_fun}
\end{aligned}
\end{equation}
for $i \in \{1,2\}$. If $\boldsymbol{\alpha}^*$ is the equilibrium policy pair, it satisfies
\begin{equation}
    J_i(\textbf{x}_i, t, \boldsymbol{\alpha}_i^*) \leq J_i(\textbf{x}_i, t, (\alpha_i,\alpha_{-i}^*)), ~\forall \alpha_{i} \in \mathcal{A},~\forall i\in \{1,2\}.
\vspace{-0.05in}
\end{equation}
Then Player $i$'s equilibrium value is $\vartheta_i(\textbf{x}_i, t) = J_i(\textbf{x}_i, t, \boldsymbol{\alpha}_i^*)$, which is the viscosity solutions to HJI equations ($L$), and satisfies the boundary condition ($D$)~\cite{starr1969nonzero}:
\begin{equation}
\begin{aligned}
& L(\vartheta_i, \nabla_{\textbf{x}_i} \vartheta_i, \textbf{x}_i, t) := \nabla_t \vartheta_i + \max_{u \in \mathcal{U}_i} \left\{\nabla_{\textbf{x}_i} \vartheta_i ^T \textbf{f}_i - (l_i + c_i)\right\} = 0 \\
& D(\vartheta_i, \textbf{x}_i) := \vartheta_i(\textbf{x}_i, T) - g_i = 0, \quad \forall~ i = 1, 2. \\
\end{aligned}
\label{eq:hji}
\end{equation}
Solving Eq.~\eqref{eq:hji} for $(\vartheta_1, \vartheta_2)$ allows us to derive Player $i$'s equilibrium policy $\alpha_i^*(\textbf{x}_i, t) = \argmax_{u \in \mathcal{U}_i} \{\nabla_{\textbf{x}_i} \vartheta_i^T \textbf{f}_i - (l_i + c_i) \}$~\cite{bressan2010noncooperative}. For player-type configuration settings, we define the Eq.\eqref{eq:hji} parameterized by $\boldsymbol{\theta}$ as $\mathcal{L}^{\boldsymbol{\theta}} := (L^{\boldsymbol{\theta}}, D^{\boldsymbol{\theta}})$. 

\textbf{Pontryagin Maximum Principle (PMP).} Although HJI equations yield closed-loop policies, solving such PDEs still has computational challenges using numerical approaches for problems beyond six-dimensional~\cite{bui2022optimizeddp}. Prior work~\cite{zhang2024value} empirically demonstrates that PMP-derived values are consistent with the ones governed by HJI. Consequently, we collect open-loop equilibrium trajectories governed by PMP and consider the formulation as follows: For an initial state $(\bar{x}_1, \bar{x}_2) \in \mathcal{X}$ at time $t \in [0, T]$, PMP satisfies
\vspace{-0.05in}
\begin{equation}
\begin{aligned}
& \dot{x}_i = f_i, \quad x_i(t) = \bar{x}_i, \\
& \dot{\lambda}_i = - \nabla_{x_i} (\lambda_i^T \textbf{f}_i - (l_i+c_i)), \quad \lambda_i(T) = - \nabla_{x_i} g_i, \\
& u_i = \argmax_{u \in \mathcal{U}_i} ~\{\lambda_i^T\textbf{f}_i - (l_i+c_i)\},  \quad \forall~ i = 1, 2.
\end{aligned}
\label{eq:pmp}
\vspace{-0.05in}
\end{equation}
where $\lambda_i = \nabla_{x_i} \vartheta_i$ is the costate of Player $i$. 
Solving Eq.~\eqref{eq:pmp} for given initial states in $\mathcal{X}$ at time $t \in [0, T]$ is a boundary value problem (BVP). 

\begin{figure}[!ht]
\centering
\includegraphics[width=1\linewidth]{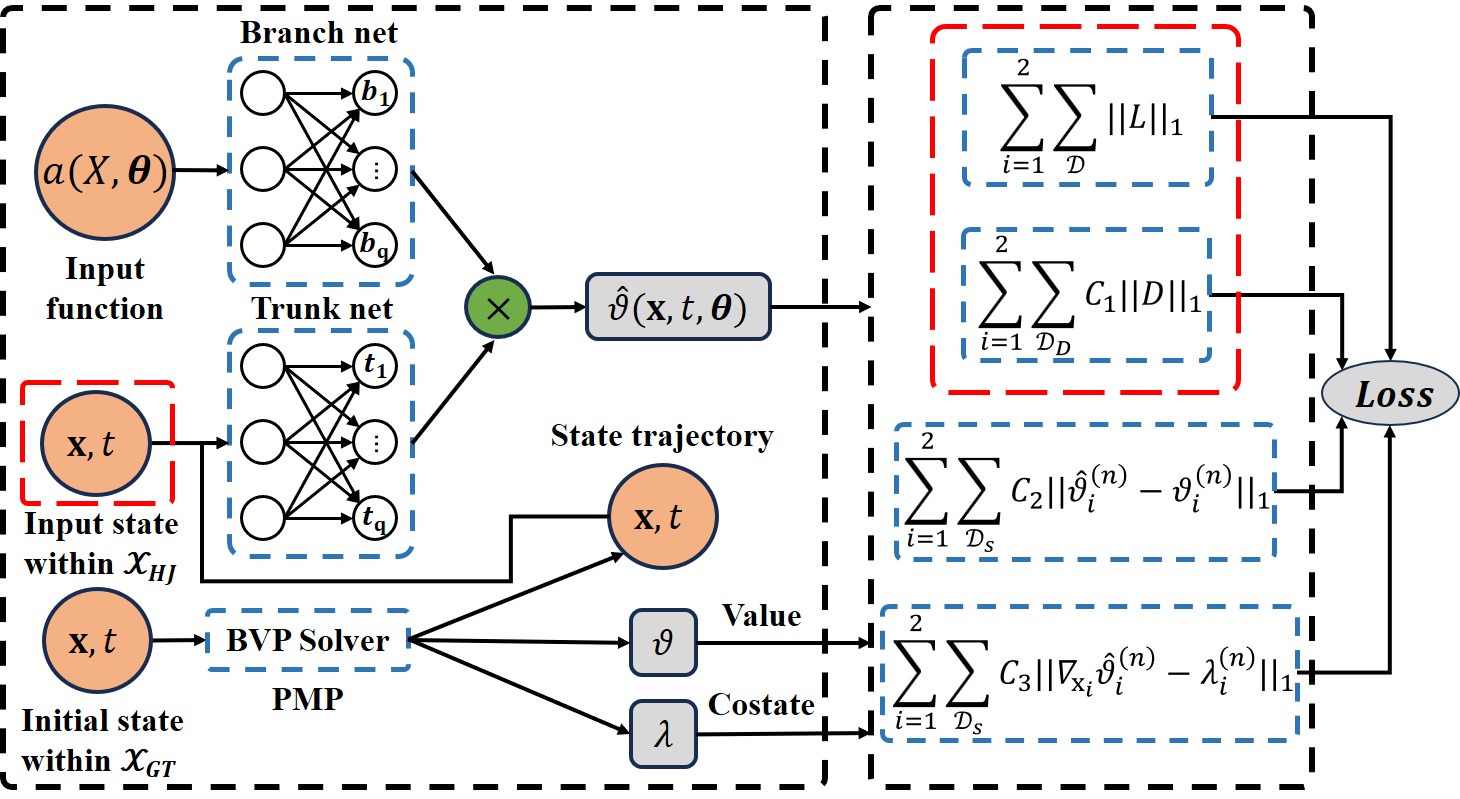}
\caption{Illustration of Hybrid Neural Operator. The baseline SNO is similar to HNO, but without red dashed box for input data points and loss function terms.}
\label{fig:hydrid Onet}
\vspace{-10pt}
\end{figure}

\textbf{Neural Operator Architecture.} We follow the classical neural operator framework and define HJI neural operator as $\hat{\vartheta}(\textbf{x}_i,t,\boldsymbol{\theta}): \mathcal{X} \times [0,T] \times \Theta^2 \rightarrow \mathbb{R}$, which maps the player-type parameters ${\boldsymbol{\theta}} \in \Theta^2$ to the values solved by $\mathcal{L}^{\boldsymbol{\theta}}$. To encode state constraints for parameter settings, we introduce an input function $a(\textbf{x},\boldsymbol{\theta}): \mathcal{X} \times \Theta^2 \rightarrow \{0,1\}$, defined as $a = 1$ if $\textbf{x}$ violates the constraints according to $\boldsymbol{\theta}$, or otherwise $a = 0$. We denote $X \in \mathbb{R}^{L \times d_x}$ as a lattice of $\mathcal{X}$ and then let $a(X, \boldsymbol{\theta}) \in \{0,1\}^{1 \times L}$ be a Boolean row vector, where each entry indicates whether a corresponding lattice node violates the state constraints. The HJI neural operator $\hat{\vartheta}$ is represented as a linear combination of basis functions:
\begin{equation}
\hat{\vartheta}(\textbf{x}, t,\boldsymbol{\theta}) = \sum_{k=1}^q \underbrace{b_k(a(X, \boldsymbol{\theta}))}_{\rm branch} \underbrace{t_k(\textbf{x}, t)}_{\rm trunk},
\label{eq:DeepONet}
\end{equation}
where branch network $b_k: \{0,1\}^{1 \times L} \rightarrow \mathbb{R}$ outputs function coefficients using PDE parameters, and trunk network $t_k: \mathbb{R}^{d_x} \times [0,T] \rightarrow \mathbb{R}$ computes basis function using input states and time. We consider two neural operator versions, HNO and SNO. The architecture of HNO is illustrated in Fig.~\ref{fig:hydrid Onet}, while SNO follows the same architecture except for the red dashed box in Fig.~\ref{fig:hydrid Onet}. We define PINN dataset for HJI equations ($L$) and boundary condition ($D$) losses as $\mathcal{D}:=\{(\textbf{x}_i,t)^{(n)} \in \mathcal{X} \times [0,T] \text{ for } i=1, 2\}_{n=1}^{N_L}$ and $\mathcal{D}_D:=\{\textbf{x}_i^{(n)} \in \mathcal{X} \text{ for } i=1, 2\}_{n=1}^{N_D}$. We also denote supervised dataset $\mathcal{D}_S:=\{(\textbf{x}_i, t, {\vartheta}_i, \nabla_{\textbf{x}_i}{\vartheta}_i)^{(n)} \text{ for } i=1, 2\}_{n=1}^{N_S}$ solved by Eq.~\eqref{eq:pmp} with initial states uniformly sampled in $\mathcal{X}$. The loss function for HNO including PINN and supervised losses is formulated as follows:
\begin{equation}
\begin{aligned}
    L_{HNO}(\hat{\boldsymbol{\vartheta}}) := 
    & \sum_{i=1}^2 \sum_{\mathcal{D}} \left\|L(\hat{{\vartheta}}_i^{(n)}, \nabla_{\textbf{x}_i} \hat{\vartheta}_i^{(n)}, \textbf{x}_i^{(n)}, t^{(n)}) \right\|_1 \\ 
    & + \sum_{\mathcal{D}_D} C_1 \left\|D(\hat{\vartheta}_i^{(n)}, \textbf{x}_i^{(n)}) \right\|_1 \\
    & + \sum_{\mathcal{D}_S} C_2\left \|\hat{\vartheta}_i^{(n)} - \vartheta_i^{(n)} \right \|_1 \\
    & + \sum_{\mathcal{D}_S} C_3 \left \|\nabla_{\textbf{x}_i}\hat{\vartheta}_i^{(n)} - {\lambda}_i^{(n)} \right\|_1.
\end{aligned}
\label{eq:loss_HNO}
\end{equation}
where $\hat{\vartheta}_i^{(n)}$ is an abbreviation for $\hat{\vartheta}(\textbf{x}_i^{(n)}, t^{(n)}, \boldsymbol{\theta}_i^{(n)})$. Hyperparameters $C_i > 0$ for $i=1,2,3$ aims to balance each loss term. For SNO, its loss function only includes the third and fourth terms formulated in Eq.~\eqref{eq:loss_HNO}.

\section{Experiments and Results}
\label{sec: experiments}
We systematically investigate the efficacy of HNO in predicted closed-loop trajectories and evaluate safety performance across three high-dimensional nonlinear case studies. The first two (narrow road avoidance and double-lane change) are formulated as general-sum games with a 9D state space, similar to previous studies on zero-sum games in~\cite{deepreach} and optimal control settings in~\cite{leung2020infusing}. These scenarios allow us to evaluate the ability of HNO to handle different two-player interactions. The third case (two-drone collision avoidance) introduces a more complex system with 13D states, representing the existing highest-dimensional HJ problem. To highlight the advantages of HNO, we use SNO based on the DeepONet architecture~\cite{lu2021learning} as a baseline.

\textbf{Hardware.} We collect supervised equilibrium data using a GTX TITAN X (12 GB memory) and train both HNO and SNO on an A100 GPU (40 GB memory).

\begin{table}[!ht]
\vspace{-5pt}
\scriptsize
\centering
\caption{Computational costs for all neural operators in all case studies}
\label{table:time cost}
\begin{tabular}{cccc}
\toprule
\multirow{1}{*}{Case Study} &
\multirow{1}{*}{Computational Cost} &
\multirow{1}{*}{Hybrid} &
\multirow{1}{*}{Supervised}\\ 
No. & (minutes) &  Neural Operator &  Neural Operator \\ \midrule
        & Data Acquisition & 333 & 667  \\
Case 1  & Neural Operator Training & 1094 & 776  \\
        & Total Time Cost & 1427 & 1443 \\
\midrule
        & Data Acquisition & 333 & 667 \\
Case 2  & Neural Operator Training & 1400 & 1092 \\
        & Total Time Cost & 1733 & 1759 \\
\midrule
        & Data Acquisition & 583 & 1167 \\
Case 3  & Neural Operator Training & 1392 & 827 \\
        & Total Time Cost & 1975 & 1994 \\
\bottomrule
\end{tabular}
\vspace{-8pt}
\end{table}

\textbf{Data.} 
To ensure a fair comparison of the learned neural operators, we consider the same computational cost including data acquisition and training, and use the same number of data points for training. Specifically, we uniformly sample initial states across the defined state space to generate 1k and 2k ground truth trajectories for HNO and SNO, respectively. Additionally, HNO samples the remaining states across the entire state space to satisfy the requirements of total training data points for comparison. We provide detailed descriptions of the data sampling in the following section and summarize the computational costs for each case study in Table~\ref{table:time cost}. For better training convergence, we normalize all input data within $[-1, 1]$.

\textbf{Network architecture and training.} Both HNO and SNO employ fully connected networks with 3 hidden layers of 64 neurons and \texttt{tanh} activation. The Adam optimizer uses a fixed learning rate of $2 \times 10^{-5}$ for Case 1 and $1 \times 10^{-4}$ for Cases 2 and 3. For HNO, we first pre-train the network for 100k iterations using supervised data. We then use curriculum learning~\cite{deepreach, krishnapriyan2021characterizing} to refine the network for an additional 200k iterations, adding additional states sampled from an expanding time window, which starts from the terminal time. In contrast, SNO refines the model with 200k iterations. To improve training efficiency, both HNO and SNO integrate adaptive activation functions~\cite{jagtap2020adaptive}. We train the model using four player-type parameters $(\theta_1, \theta_2)=\{(1, 1), (1, 5), (5, 1), (5, 5)\}$ and evaluate the model generalization performance within the parameter space $\Theta$ for three case studies. 

All three case studies use state constraints to prevent collisions during two-player interactions. Therefore, our model performance analysis focuses on collision rate (Col.\%) as the evaluation metric. The collision rate represents the deviation from theoretical safety, which is zero when using ground-truth solutions computed through BVP solvers. However, collisions may occur due to value approximation errors in neural operators. We define the collision rate as $Col.\% = N_{pred}/N_{gt}$, where $N_{pred}$ is the number of trajectories that lead to collisions when using the neural operator as closed-loop controllers and $N_{gt}$ is the number of collision-free trajectories solved by BVP. Neural operators use the same uniformly sampled initial states as BVP solvers during simulations. 

We propose the following hypothesis: 
\textit{Given the same computational budget, the Hybrid Neural Operator (HNO) achieves high safety performance (low collision rates) compared to the Supervised Neural Operator (SNO) across the entire parameter space in all case studies}.

\subsection{Case 1: narrow road collision avoidance} 
\begin{figure}[!ht]
\vspace{-5pt}
\centering
\includegraphics[width=0.8\linewidth]{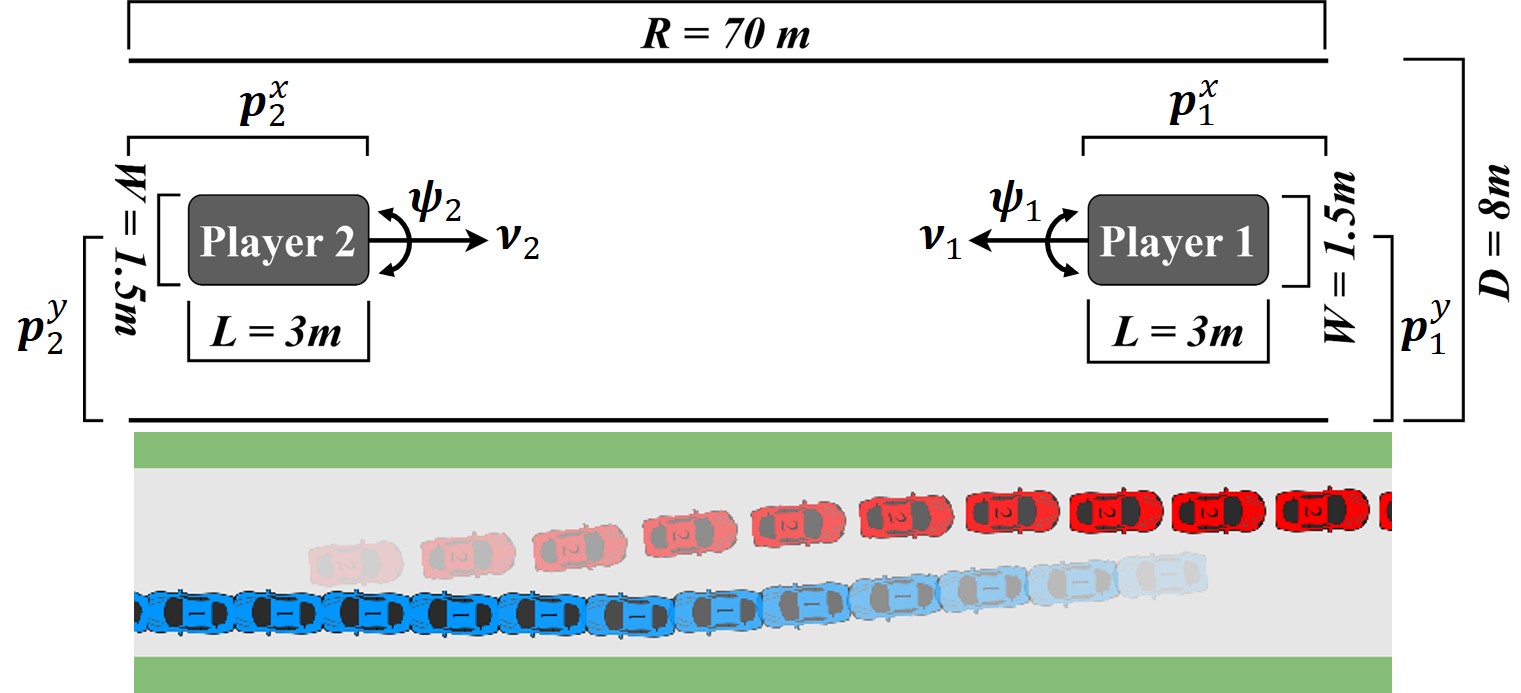}
\caption{Narrow road collision avoidance scenario. Simulation shows the ground truth trajectory for player-type configuration $(\theta_1,\theta_2)=(1,1)$.}
\label{fig:narrow road case}
\vspace{-5pt}
\end{figure}

We first consider Case 1, illustrated in Fig.~\ref{fig:narrow road case}, where each Player $i$ is characterized by its position ($p^{x}_i$, $p^{y}_i$), orientation ($\psi_i$), and speed ($v_i$), forming the state vector $x_i := [p^{x}_i, p^{y}_i, \psi_i, v_i]^T$. The system uses a unicycle dynamics model:
\begin{eqnarray}
\begin{aligned}
\left[
\begin{array}{c}
    \dot p^{x}_i \\
    \dot p^{y}_i \\
    \dot \psi_i \\
    \dot v_i \\
\end{array}
\right]
=
\left[
\begin{array}{c}
    v_i\cos(\psi_i) \\
    v_i\sin(\psi_i) \\
    \omega_i \\
    u_i \\
\end{array}
\right],
\end{aligned}
\end{eqnarray} 
where $\omega_i \in [-1, 1] rad/s$ and $u_i \in [-5, 10] m/s^2$ represent the angular velocity and acceleration control inputs, respectively. The instantaneous loss and state constraint functions are defined as follows:
\begin{equation}
\begin{aligned}
    & l_i^{\boldsymbol{\theta}}(x_i, u_i) = k\omega_i^2 + u_i^2, \\
    & c_i^{\boldsymbol{\theta}}(\textbf{x}_i) = b\left(1+\exp(-\gamma (\eta^{\boldsymbol{\theta}}(\textbf{x}_i)-S)\right))^{-1}, 
\end{aligned}
\label{eq:case 1_instantenous loss}
\end{equation}
where $k=100$, $b=10^4$ (parameter for a high penalty on collisions), and $\gamma=5$ (shape parameter). The distance function between two players is defined as $S = \sqrt{((R - p^{x}_2) - p^{x}_1)^2 + (p^{y}_2 - p^{y}_1)^2}$, where $R=70 m$. The collision threshold $\eta^{\boldsymbol{\theta}}(\textbf{x}_i)$ is defined as:
\begin{equation}
\begin{aligned}
    \eta^{\boldsymbol{\theta}}(\textbf{x}_i) = 0.1(\theta_1+\theta_2)+0.05\min(\theta_1, \theta_2)+1.25.
\end{aligned}
\end{equation}
The terminal loss function encourages players to stay in their lanes while maintaining nominal speed:
\begin{equation}
g_i(x_i) = -\mu p^x_{i}(T) + (v_{i}(T)-\bar{v})^2 + (p^y_{i}(T)-\bar{p}^{y})^2,
\label{eq:case 1_terminal loss}
\end{equation}
where $\mu = 10^{-6}$, $\bar{v} = 18 m/s$, $\bar{p}^{y} = 4 m$, and $T = 3s$.

We generate 1k and 2k ground truth trajectories for HNO and SNO, respectively, by uniformly sampling initial states from $\mathcal{X}_{GT}:= [15, 20]m \times [3.25, 4.75]m \times [-\pi/180, \pi/180] rad \times [18, 25]m/s$ for HNO and SNO, respectively. Each trajectory involves 31 data points with a time interval of 0.1s, resulting in 62k and 124k data points for HNO and SNO. Additionally, 62k states are sampled from $\mathcal{X}_{HJ}:= [15, 95]m \times [0, 8]m \times [-0.2, 0.2]rad \times [18, 29]m/s$ for training data points to refine HNO.

\subsection{Case 2: double-lane change} 
\begin{figure}[!ht]
\centering
\includegraphics[width=0.9\linewidth]{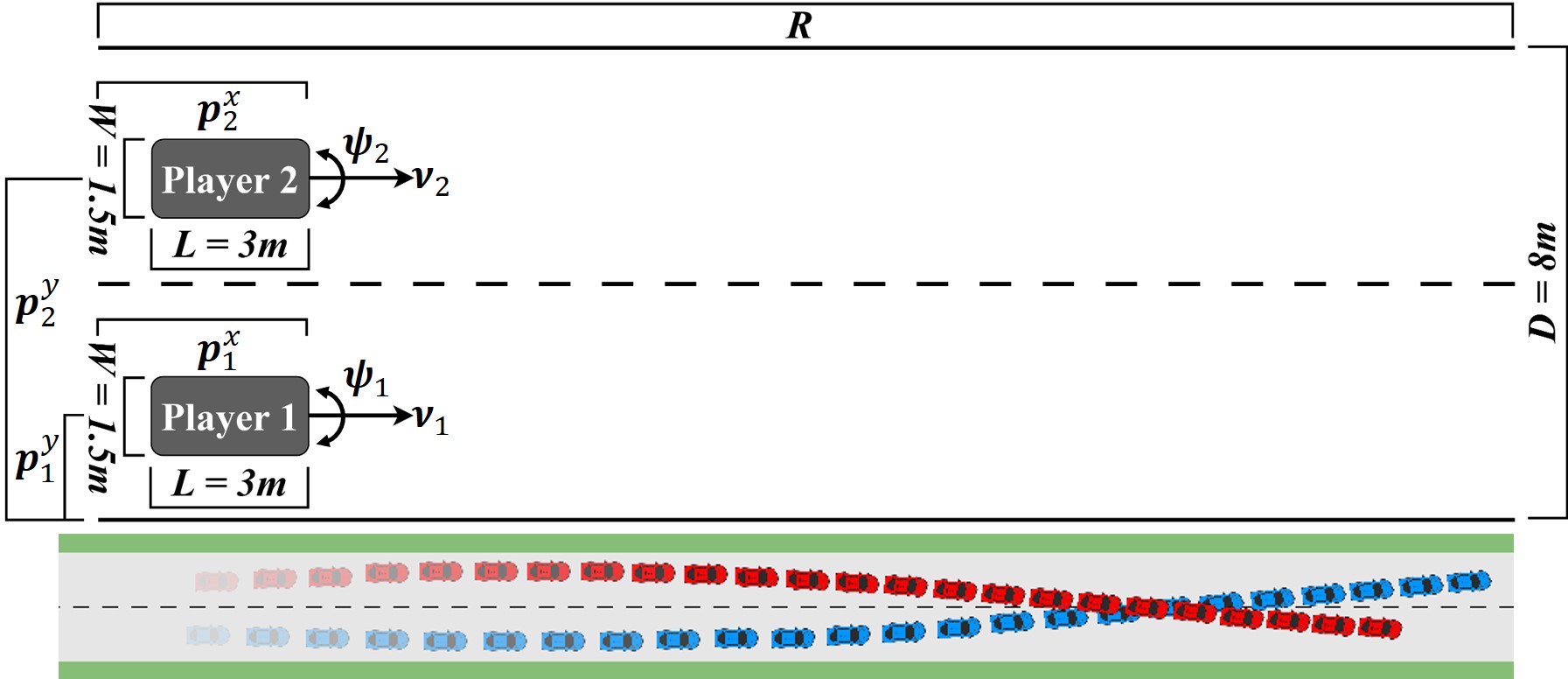}
\caption{Double-lane change scenario. Simulation shows the ground truth trajectory for player-type configuration $(\theta_1,\theta_2)=(1,1)$.}
\label{fig:lane change case}
\vspace{-15pt}
\end{figure}

The scenario for Case 2 is depicted in Fig.~\ref{fig:lane change case}. The unicycle dynamics, instantaneous loss, and state constraints remain the same as in Case 1, with a modified distance function $S = \sqrt{((p^{x}_2 - p^{x}_1)^2 + (p^{y}_2 - p^{y}_1)^2}$. Additionally, we consider the collision threshold $\eta^{\boldsymbol{\theta}}(\textbf{x}_i)$ as:
\begin{equation}
\begin{aligned}
    \eta^{\boldsymbol{\theta}}(\textbf{x}_i) = 0.1(\theta_1+\theta_2)+0.05\min(\theta_1, \theta_2)+2.25.
\end{aligned}
\end{equation}
The terminal loss function ensures lane adherence and nominal speed restoration:
\begin{equation}
\begin{aligned}
g_i(x_i) = -\mu p^x_{i}(T) + (p^y_{i}(T)-\bar{p}^{y}_i)^2 +  \\
(v_{i}(T)-\bar{v})^2 +k_{\psi}(\psi_{i}(T)-\bar{\psi})^2,
\end{aligned}
\label{eq:case 2_terminal loss}
\end{equation}
where $\mu = 10^{-6}$, $k_{\psi} = 100$, $\bar{p}^{y}_1 = 6 m$ for player 1 and $\bar{p}^{y}_2 = 2 m$ for player 2, $\bar{v} = 18 m/s$, $\bar{\psi} = 0 rad$, and $T = 4s$.

We generate supervised data by uniformly sampling initial states from $\mathcal{X}_{GT}^1:= [0, 3]m \times [1.25, 2.75]m \times [-\pi/180, \pi/180] rad \times [18, 25]m/s$ for play 1, and $\mathcal{X}_{GT}^2:= [0, 3]m \times [5.25, 6.75]m \times [-\pi/180, \pi/180] rad \times [18, 25]m/s$ for play 2. Consequently, HNO and SNO collect 1k and 2k ground truth trajectories, respectively. Each trajectory contains 41 data points with a 0.1s time interval, yielding a total of 82k and 164k data points for HNO and SNO. Additionally, for HNO refinement, we uniformly sample 82k states from $\mathcal{X}_{HJ}^1 = [0, 90]m \times [0, 6]m \times [-0.17, 0.15]rad \times [17, 26]m/s$ for player 1 and $\mathcal{X}_{HJ}^2 = [0, 90]m \times [2, 8]m \times [-0.15, 0.17]rad \times [17, 26]m/s$ for player 2.

\subsection{Case 3: two-drone collision avoidance}
\begin{figure}[!ht]
\centering
\includegraphics[width=0.7\linewidth]{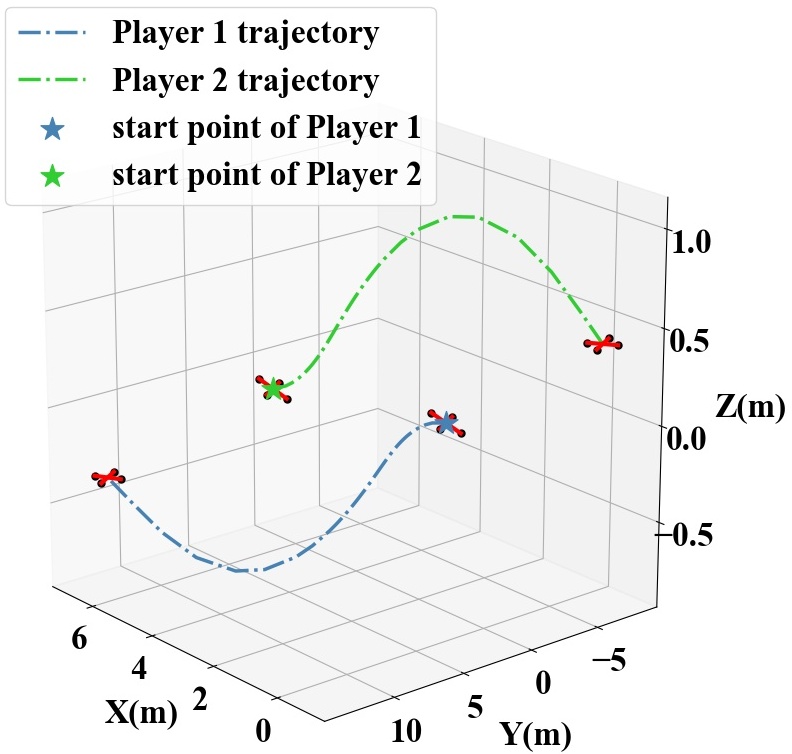}
\caption{Ground truth trajectory of two-drone collision avoidance for player-type configuration $(\theta_1,\theta_2)=(1,1)$.}
\label{fig:drone case}
\vspace{-18pt}
\end{figure}

Lastly, we consider Case 3, a 13D nonlinear model based on drone flight dynamics (see ground truth trajectory in Fig.~\ref{fig:drone case}), assuming zero yaw angle with respect to a global coordinate frame. The state of Player $i$ consists of its location ($p^{x}_i$, $p^{y}_i$, $p^{z}_i$), and speed ($v^{x}_i$, $v^{y}_i$, $v^{z}_i$), forming the state vector $x_i := [p^{x}_i, p^{y}_i, p^{z}_i, v^{x}_i, v^{y}_i, v^{z}_i]^T$. The system uses the drone dynamics model described in~\cite{fridovich2020confidence}: 
\begin{eqnarray}
\begin{aligned}
\left[
\begin{array}{c}
    \dot p_i^x \\
    \dot p_i^y \\
    \dot p_i^z \\
    \dot v_i^x \\
    \dot v_i^y \\
    \dot v_i^z \\
\end{array}
\right]
=
\left[
\begin{array}{c}
    v_i^x \\
    v_i^y \\
    v_i^z \\
    g \tan(\psi_i) \\
    -g \tan(\phi_i) \\
    \tau_i - g \\
\end{array}
\right],
\end{aligned}
\end{eqnarray} 
where $u_i = (\psi_i, \phi_i, \tau_i)$ represents roll, pitch, and thrust controls. The control inputs satisfy $\psi_i \in [-0.05, 0.05] rad$, $\phi_i \in [-0.05, 0.05] rad$, $\tau_i \in [7.81, 11.81] m/s^2$ with $g = 9.81 m/s^2$. We give the instantaneous loss and state constraints as:
\begin{equation}
\begin{aligned}
    & l_i^{\boldsymbol{\theta}}(x_i, \omega_i, u_i) = k_{\psi}\tan^2(\psi_i) + k_{\phi}\tan^2(\phi_i) + (\tau_i - g)^2, \\
    & c_i^{\boldsymbol{\theta}}(\textbf{x}_i) = b\left(1+\exp(-\gamma (\eta^{\boldsymbol{\theta}}(\textbf{x}_i)-S)\right))^{-1}, 
\end{aligned}
\end{equation}
where $b = 10^4$, $\gamma=5$. $k_{\psi}=k_{\phi}=100$, balancing control efforts for roll, pitch, and thrust. The distance function between the two players is defined as:
\begin{equation}
\begin{aligned}
    & S = \sqrt{((R_x - p_2^x) - p_1^x)^2  + ((R_y - p_2^y) - p_1^y)^2 + (p_2^z - p_1^z)^2}. \notag
\end{aligned}
\end{equation}
where $R_x=5m$ and $R_y=5m$ shift the players’ coordinates along the $x$- and $y$-axes. Furthermore, the collision threshold $\eta^{\boldsymbol{\theta}}(\textbf{x}_i)$ is defined as:
\begin{equation}
\begin{aligned}
    \eta^{\boldsymbol{\theta}}(\textbf{x}_i) = 0.1(\theta_1+\theta_2)+0.05\min(\theta_1, \theta_2)+0.5.
\end{aligned}
\end{equation}
The terminal loss encourages players to stabilize their positions and velocities when the game for two players is complete:
\begin{equation}
\begin{aligned}
g_i(x_i) = -\mu p^x_i(T) - \mu p^y_i(T) + (p^z_{i}(T)-\bar{p}^{z}_i)^2 +  \\
(v^x_i(T) - \bar{v}^x_i)^2 + (v^y_i(T) - \bar{v}^y_i)^2 + (v^z_i(T) - \bar{v}^z_i)^2.
\end{aligned}
\end{equation}
where $\mu = 10^{-6}$, $\bar{p}^z_i = 0 m$ , $\bar{v}^x_i = \bar{v}^y_i = \bar{v}^z_i = 0 m/s$, and $T = 4s$. 

We generate 1k and 2k ground truth trajectories by uniformly sampling initial states from $\mathcal{X}_{GT}:= [0, 1]m \times [0, 1]m \times [-0.1, 0.1]m \times [2, 4]m/s \times [2, 4]m/s \times [0, 0.1]m/s$ for HNO and SNO, respectively. Each trajectory involves 41 data points with a 0.1s time interval, leading to a total of 82k and 164k data points for HNO and SNO. Additionally, for HNO refinement, we sample 82k states uniformly from $\mathcal{X}_{HJ}:= [0, 15.5]m \times [0, 15.5]m \times [-2.2, 2.5]m \times [0.3, 4.5]m/s \times [0.3, 4.5]m/s \times [-2, 2.2]m/s$.


\begin{figure}[!ht]
\vspace{-10pt}
\centering
\includegraphics[width=1\linewidth]{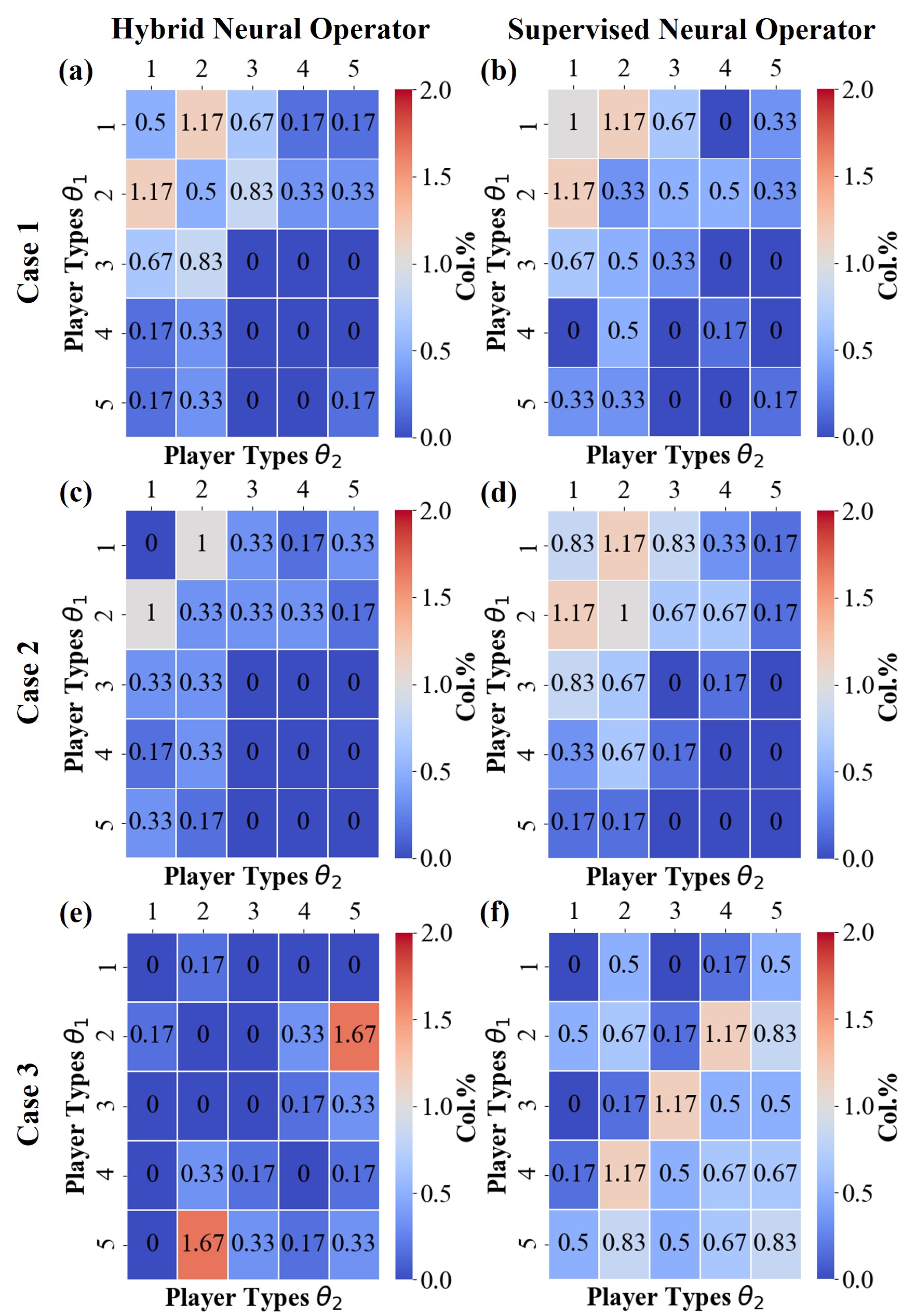}
\caption{Safety performance (Col.\%) comparison among HNO and SNO for each parameter configuration in $\Theta^2$ across all case studies.}
\label{fig:heatmap}
\vspace{-10pt}
\end{figure}

\subsection{Experimental Results Discussion} 
\textbf{Safety performance.} 
For each case study, we collect 600 test collision-free trajectories (zero collision rates), with initial states uniformly sampled from $\mathcal{X}_{GT}$ by solving Eq.~\eqref{eq:pmp}. Starting from these initial states, we generate the trajectories by applying $\nabla_{\textbf{x}_i} \hat\vartheta_i, \forall i=1,2$ as the closed-loop control for both HNO and SNO. We then compute the collision rate (Col.\%) for both models and summarize the results in the heatmap (see Fig.~\ref{fig:heatmap}). It is first noted that HNO demonstrates lower collision rates compared to SNO for trained player-type parameters ($(\theta_1, \theta_2) = {(1,1), (1,5), (5,1), (5,5)}$) under the same computational budget. For unseen player-type parameters, HNO also outperforms SNO and has low collision rates across all case studies. We also notice that SNO has high collision rates for most player-type parameters, especially in the higher dimensional case study (Case 3). The reason for the better performance of HNO is that it leverages informative supervised data to capture the landscape for values and their gradients, and simultaneously samples $(\textbf{x}, t)$ across the entire spatial-temporal domain to satisfy PDE constraints. In contrast, SNO only learns from fixed trajectories. Our current experimental results for HNO are consistent with previous studies~\cite{zhang2024value}, which also demonstrate the advantage of combining supervised learning and PINN approach.


\begin{table}[!ht]
\vspace{-5pt}
\scriptsize
\centering
\caption{Ratio of NTK condition numbers with different activations}
\label{table:condition_number}
\begin{tabular}{ccc}
\toprule
\multirow{1}{*}{Player-type} &
\multicolumn{2}{c}{Ratio of NTK Condition Numbers} \\ 
\cmidrule(lr){2-3} 
Parameters & $r_1 = \kappa(\texttt{tanh}) / \kappa(\texttt{sin})$ &  $r_2 = \kappa(\texttt{tanh}) / \kappa(\texttt{relu})$ \\ \midrule
(1,1)   & $1.97 \times 10^{-3}$ & $1.40 \times 10^{-10}$ \\
\midrule
(1,5)  & $4.53 \times 10^{-3}$ & $1.43 \times 10^{-10}$  \\
\midrule
(5,1)  & $5.98 \times 10^{-5}$ & $2.37 \times 10^{-9}$  \\
\midrule
(5,5)  & $1.98 \times 10^{-4}$ & $1.41 \times 10^{-8}$ \\
\bottomrule
\end{tabular}
\vspace{-5pt}
\end{table}

\begin{figure}[!ht]
\vspace{-10pt}
\centering
\includegraphics[width=1\linewidth]{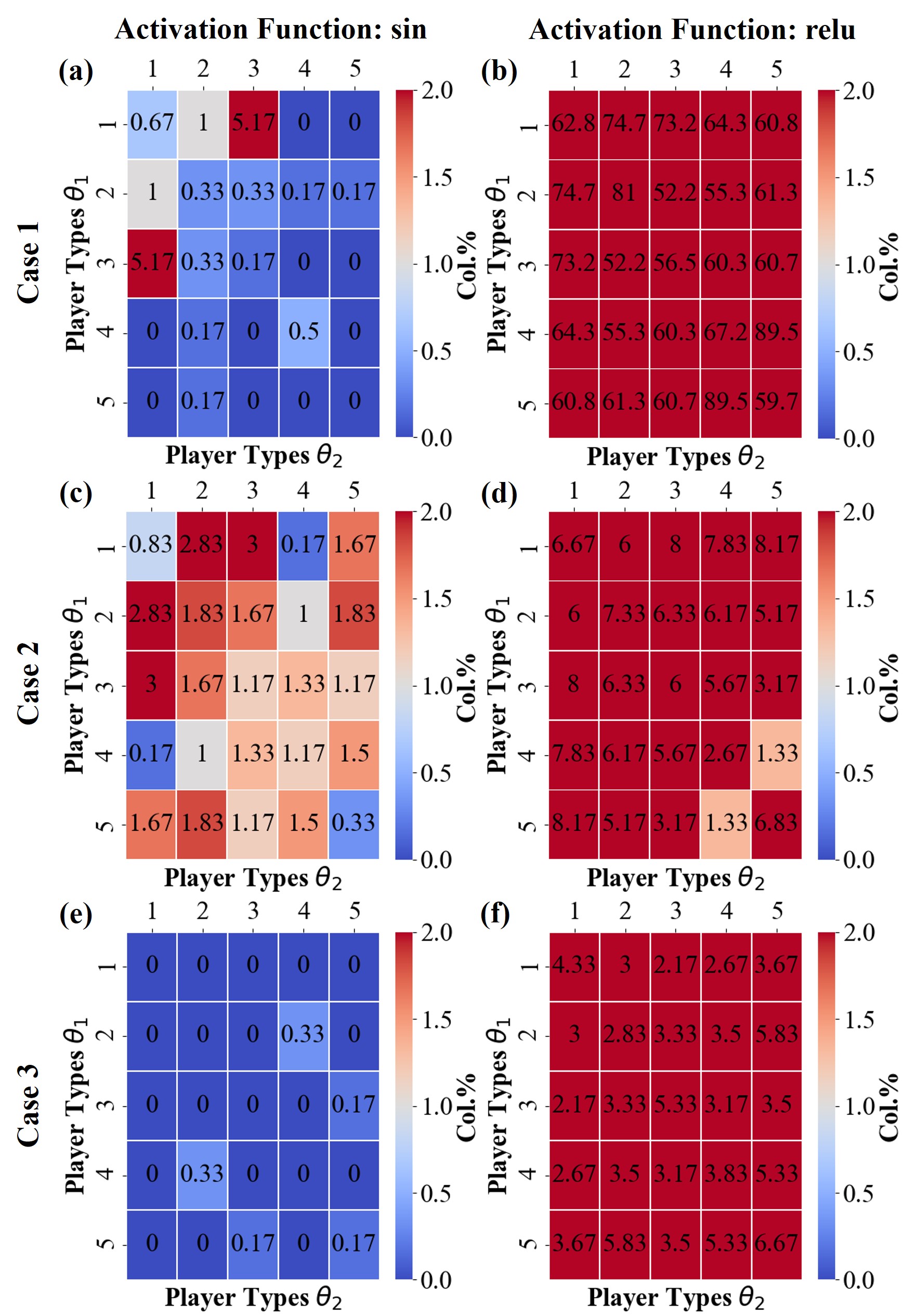}
\caption{Safety performance (Col.\%) comparison among \texttt{sin} and \texttt{relu} using HNO for each parameter configuration in $\Theta^2$ across all case studies.}
\label{fig:activation_choice}
\vspace{-10pt}
\end{figure}

\textbf{Ablation studies.} 
Several recent studies highlight the importance of activation function choice in physics-informed model performance~\cite{deepreach,zhang2024value}. To further investigate this effect, we conduct ablation studies to evaluate the efficacy of different activation functions for HNO in high-dimensional nonlinear dynamics. Fig.~\ref{fig:activation_choice} indicates that \texttt{tanh} has good stability and high safety performance compared to \texttt{sin} and \texttt{relu} across the parameter space in all case studies. To have a straightforward insight into why activation choice affects model performance, we analyze Case 2, where the collision rate differences among activation functions are most obvious. Specifically, we compute the neural tangent kernel (NTK) for HNO using \texttt{tanh}, \texttt{sin}, and \texttt{relu}, respectively. Recent studies~\cite{jacot2018neural} show that an infinitely wide neural network is a kernel machine. This kernel machine, which is called NTK, is determined by the network architecture~\cite{jacot2018neural, lee2019wide}. Existing NTK analysis for PINNs~\cite{wang2022and} demonstrates that model convergence is related to the eigenvalues and eigenvectors of the NTK. In particular, a well-conditioned NTK, characterized by a low condition number, is directly correlated with stable training and better generalization~\cite{lau2024pinnacle, mohamadi2023fast}. Based on this insight, we compute condition number $\kappa=\lambda_{max}/\lambda_{min}$ for each activation function using training data across the four player-type parameters $(\theta_1, \theta_2) = \{(1,1), (1,5), (5,1), (5,5)\}$, and report the results in Table~\ref{table:condition_number}. The results show that $r_1 < 1$ and $r_2 < 1$, which indicates that \texttt{tanh} outperforms \texttt{sin} and \texttt{relu} in terms of generalization performance due to low condition number. While NTK is a useful theoretical framework to analyze model generalization, finite-width networks often have better generalization properties compared to infinitely wide ones. Consequently, not all observed phenomena in neural networks can be fully explained through NTK analysis alone. In the future, we will further investigate the activation function choice in terms of safety and generalization performance using NTK analysis.


\section{Conclusion}
We propose HNO for parametric discontinuous value approximation in two-player general-sum differential games with state constraints. HNO extends DeepONet by integrating supervised data and physics-informed constraints, addressing the challenges of discontinuous value functions caused by state constraints across the parameter space. Empirical results demonstrate that HNO achieves high safety performance compared to the baseline across high-dimensional nonlinear case studies under the same computational cost. However, HNO still relies on supervised data generated via BVPs, which can suffer from convergence issues due to multiple equilibria and singular arcs~\cite{singarcs}. In future work, we will explore the epigraphical techniques~\cite{altarovici2013general,lee2021hamilton,zhang2024value} to develop a fully self-supervised approach for parametric value approximation. Additionally, to ensure a more comprehensive evaluation, we will extend our comparisons to include PNO~\cite{zhang2024pontryagin} and constrained reinforcement learning methods (e.g., RC-PPO~\cite{so2025solving}) to further assess the efficacy of HNO in differential games.

\bibliography{iros2025}

\begin{thebibliography}{10}
\providecommand{\url}[1]{#1}
\csname url@samestyle\endcsname
\providecommand{\newblock}{\relax}
\providecommand{\bibinfo}[2]{#2}
\providecommand{\BIBentrySTDinterwordspacing}{\spaceskip=0pt\relax}
\providecommand{\BIBentryALTinterwordstretchfactor}{4}
\providecommand{\BIBentryALTinterwordspacing}{\spaceskip=\fontdimen2\font plus
\BIBentryALTinterwordstretchfactor\fontdimen3\font minus \fontdimen4\font\relax}
\providecommand{\BIBforeignlanguage}[2]{{%
\expandafter\ifx\csname l@#1\endcsname\relax
\typeout{** WARNING: IEEEtran.bst: No hyphenation pattern has been}%
\typeout{** loaded for the language `#1'. Using the pattern for}%
\typeout{** the default language instead.}%
\else
\language=\csname l@#1\endcsname
\fi
#2}}
\providecommand{\BIBdecl}{\relax}
\BIBdecl

\bibitem{spica2020real}
R.~Spica, E.~Cristofalo, Z.~Wang, E.~Montijano, and M.~Schwager, ``A real-time game theoretic planner for autonomous two-player drone racing,'' \emph{IEEE Transactions on Robotics}, vol.~36, no.~5, pp. 1389--1403, 2020.

\bibitem{zhang2023approximating}
L.~Zhang, M.~Ghimire, W.~Zhang, Z.~Xu, and Y.~Ren, ``{Approximating discontinuous Nash equilibrial values of two-player general-Sum differential games},'' in \emph{2023 IEEE International Conference on Robotics and Automation (ICRA)}.\hskip 1em plus 0.5em minus 0.4em\relax IEEE, 2023, pp. 3022--3028.

\bibitem{ghimire2024state}
M.~Ghimire, L.~Zhang, Z.~Xu, and Y.~Ren, ``State-constrained zero-sum differential games with one-sided information,'' \emph{arXiv preprint arXiv:2403.02741}, 2024.

\bibitem{muchen2024mixed}
M.~Muchen~Sun, F.~Baldini, K.~Hughes, P.~Trautman, and T.~Murphey, ``Mixed strategy nash equilibrium for crowd navigation,'' \emph{The International Journal of Robotics Research}, p. 02783649241302342, 2024.

\bibitem{fridovich2020efficient}
D.~Fridovich-Keil, E.~Ratner, L.~Peters, A.~D. Dragan, and C.~J. Tomlin, ``Efficient iterative linear-quadratic approximations for nonlinear multi-player general-sum differential games,'' in \emph{2020 IEEE international conference on robotics and automation (ICRA)}.\hskip 1em plus 0.5em minus 0.4em\relax IEEE, 2020, pp. 1475--1481.

\bibitem{schwarting2019social}
W.~Schwarting, A.~Pierson, J.~Alonso-Mora, S.~Karaman, and D.~Rus, ``Social behavior for autonomous vehicles,'' \emph{Proceedings of the National Academy of Sciences}, vol. 116, no.~50, pp. 24\,972--24\,978, 2019.

\bibitem{cleac2019algames}
S.~L. Cleac'h, M.~Schwager, and Z.~Manchester, ``Algames: A fast solver for constrained dynamic games,'' \emph{arXiv preprint arXiv:1910.09713}, 2019.

\bibitem{zhang2024value}
L.~Zhang, M.~Ghimire, W.~Zhang, Z.~Xu, and Y.~Ren, ``Value approximation for two-player general-sum differential games with state constraints,'' \emph{IEEE Transactions on Robotics}, 2024.

\bibitem{viscosity}
M.~G. Crandall and P.-L. Lions, ``{Viscosity solutions of Hamilton-Jacobi equations},'' \emph{Transactions of the American mathematical society}, vol. 277, no.~1, pp. 1--42, 1983.

\bibitem{mitchell2005time}
I.~M. Mitchell, A.~M. Bayen, and C.~J. Tomlin, ``A time-dependent hamilton-jacobi formulation of reachable sets for continuous dynamic games,'' \emph{IEEE Transactions on automatic control}, vol.~50, no.~7, pp. 947--957, 2005.

\bibitem{mitchell2003}
I.~M. Mitchell and C.~J. Tomlin, ``{Overapproximating reachable sets by Hamilton-Jacobi projections},'' \emph{journal of Scientific Computing}, vol.~19, no.~1, pp. 323--346, 2003.

\bibitem{darbon2020overcoming}
J.~Darbon, G.~P. Langlois, and T.~Meng, ``Overcoming the curse of dimensionality for some hamilton--jacobi partial differential equations via neural network architectures,'' \emph{Research in the Mathematical Sciences}, vol.~7, no.~3, p.~20, 2020.

\bibitem{mukherjee2023bridging}
A.~Mukherjee and J.~Liu, ``Bridging physics-informed neural networks with reinforcement learning: Hamilton-jacobi-bellman proximal policy optimization (hjbppo),'' \emph{arXiv preprint arXiv:2302.00237}, 2023.

\bibitem{yu2022gradient}
J.~Yu, L.~Lu, X.~Meng, and G.~E. Karniadakis, ``{Gradient-enhanced physics-informed neural networks for forward and inverse PDE problems},'' \emph{Computer Methods in Applied Mechanics and Engineering}, vol. 393, p. 114823, 2022.

\bibitem{wang2021learning}
S.~Wang, H.~Wang, and P.~Perdikaris, ``{Learning the solution operator of parametric partial differential equations with physics-informed DeepONets},'' \emph{Science advances}, vol.~7, no.~40, p. eabi8605, 2021.

\bibitem{zhang2024pontryagin}
L.~Zhang, M.~Ghimire, Z.~Xu, W.~Zhang, and Y.~Ren, ``Pontryagin neural operator for solving general-sum differential games with parametric state constraints,'' in \emph{6th Annual Learning for Dynamics \& Control Conference}.\hskip 1em plus 0.5em minus 0.4em\relax PMLR, 2024, pp. 1728--1740.

\bibitem{lu2021learning}
L.~Lu, P.~Jin, G.~Pang, Z.~Zhang, and G.~E. Karniadakis, ``{Learning nonlinear operators via DeepONet based on the universal approximation theorem of operators},'' \emph{Nature machine intelligence}, vol.~3, no.~3, pp. 218--229, 2021.

\bibitem{cardaliaguet2000pursuit}
P.~Cardaliaguet, M.~Quincampoix, and P.~Saint-Pierre, ``Pursuit differential games with state constraints,'' \emph{SIAM Journal on Control and Optimization}, vol.~39, no.~5, pp. 1615--1632, 2000.

\bibitem{altarovici2013general}
A.~Altarovici, O.~Bokanowski, and H.~Zidani, ``{A general Hamilton-Jacobi framework for non-linear state-constrained control problems},'' \emph{ESAIM: Control, Optimisation and Calculus of Variations}, vol.~19, no.~2, pp. 337--357, 2013.

\bibitem{lee2021hamilton}
D.~Lee and C.~J. Tomlin, ``Hamilton-jacobi equations for two classes of state-constrained zero-sum games,'' \emph{arXiv preprint arXiv:2106.15006}, 2021.

\bibitem{gammoudi2023differential}
N.~Gammoudi and H.~Zidani, ``A differential game control problem with state constraints,'' \emph{Mathematical Control and Related Fields}, vol.~13, no.~2, pp. 554--582, 2023.

\bibitem{osher2004level}
S.~Osher, R.~Fedkiw, and K.~Piechor, ``Level set methods and dynamic implicit surfaces,'' \emph{Appl. Mech. Rev.}, vol.~57, no.~3, pp. B15--B15, 2004.

\bibitem{mitchell2005toolbox}
I.~M. Mitchell and J.~A. Templeton, ``{A toolbox of Hamilton-Jacobi solvers for analysis of nondeterministic continuous and hybrid systems},'' in \emph{Hybrid Systems: Computation and Control: 8th International Workshop, HSCC 2005, Zurich, Switzerland, March 9-11, 2005. Proceedings 8}.\hskip 1em plus 0.5em minus 0.4em\relax Springer, 2005, pp. 480--494.

\bibitem{osher1991high}
S.~Osher and C.-W. Shu, ``{High-order essentially nonoscillatory schemes for Hamilton--Jacobi equations},'' \emph{SIAM Journal on numerical analysis}, vol.~28, no.~4, pp. 907--922, 1991.

\bibitem{weinan2021algorithms}
W.~E, J.~Han, and A.~Jentzen, ``{Algorithms for solving high dimensional PDEs: from nonlinear Monte Carlo to machine learning},'' \emph{Nonlinearity}, vol.~35, no.~1, p. 278, 2021.

\bibitem{han2020convergence}
J.~Han and J.~Long, ``Convergence of the deep bsde method for coupled fbsdes,'' \emph{Probability, Uncertainty and Quantitative Risk}, vol.~5, no.~1, pp. 1--33, 2020.

\bibitem{han2018solving}
J.~Han, A.~Jentzen, and W.~E, ``Solving high-dimensional partial differential equations using deep learning,'' \emph{Proceedings of the National Academy of Sciences}, vol. 115, no.~34, pp. 8505--8510, 2018.

\bibitem{jagtap2020adaptive}
A.~D. Jagtap, K.~Kawaguchi, and G.~E. Karniadakis, ``Adaptive activation functions accelerate convergence in deep and physics-informed neural networks,'' \emph{Journal of Computational Physics}, vol. 404, p. 109136, 2020.

\bibitem{deepreach}
S.~Bansal and C.~J. Tomlin, ``{DeepReach}: A deep learning approach to high-dimensional reachability,'' in \emph{2021 IEEE International Conference on Robotics and Automation (ICRA)}.\hskip 1em plus 0.5em minus 0.4em\relax IEEE, 2021, pp. 1817--1824.

\bibitem{nakamura2021}
T.~Nakamura-Zimmerer, Q.~Gong, and W.~Kang, ``{Adaptive deep learning for high-dimensional Hamilton--Jacobi--Bellman equations},'' \emph{SIAM Journal on Scientific Computing}, vol.~43, no.~2, pp. A1221--A1247, 2021.

\bibitem{shin2020convergence}
Y.~Shin, J.~Darbon, and G.~E. Karniadakis, ``On the convergence of physics informed neural networks for linear second-order elliptic and parabolic type pdes,'' \emph{arXiv preprint arXiv:2004.01806}, 2020.

\bibitem{ito2021neural}
K.~Ito, C.~Reisinger, and Y.~Zhang, ``{A neural network-based policy iteration algorithm with global $H^{2}$-superlinear convergence for stochastic games on domains},'' \emph{Foundations of Computational Mathematics}, vol.~21, no.~2, pp. 331--374, 2021.

\bibitem{lau2024pinnacle}
G.~K.~R. Lau, A.~Hemachandra, S.-K. Ng, and B.~K.~H. Low, ``Pinnacle: Pinn adaptive collocation and experimental points selection,'' \emph{arXiv preprint arXiv:2404.07662}, 2024.

\bibitem{krishnapriyan2021characterizing}
A.~Krishnapriyan, A.~Gholami, S.~Zhe, R.~Kirby, and M.~W. Mahoney, ``Characterizing possible failure modes in physics-informed neural networks,'' \emph{Advances in Neural Information Processing Systems}, vol.~34, pp. 26\,548--26\,560, 2021.

\bibitem{mojgani2023kolmogorov}
R.~Mojgani, M.~Balajewicz, and P.~Hassanzadeh, ``{Kolmogorov n--width and Lagrangian physics-informed neural networks: a causality-conforming manifold for convection-dominated PDEs},'' \emph{Computer Methods in Applied Mechanics and Engineering}, vol. 404, p. 115810, 2023.

\bibitem{kovachki2023neural}
N.~Kovachki, Z.~Li, B.~Liu, K.~Azizzadenesheli, K.~Bhattacharya, A.~Stuart, and A.~Anandkumar, ``Neural operator: Learning maps between function spaces with applications to pdes,'' \emph{Journal of Machine Learning Research}, vol.~24, no.~89, pp. 1--97, 2023.

\bibitem{chen1995universal}
T.~Chen and H.~Chen, ``Universal approximation to nonlinear operators by neural networks with arbitrary activation functions and its application to dynamical systems,'' \emph{IEEE transactions on neural networks}, vol.~6, no.~4, pp. 911--917, 1995.

\bibitem{li2020fourier}
Z.~Li, N.~Kovachki, K.~Azizzadenesheli, B.~Liu, K.~Bhattacharya, A.~Stuart, and A.~Anandkumar, ``Fourier neural operator for parametric partial differential equations,'' \emph{arXiv preprint arXiv:2010.08895}, 2020.

\bibitem{wang2024latent}
T.~Wang and C.~Wang, ``Latent neural operator for solving forward and inverse pde problems,'' \emph{arXiv preprint arXiv:2406.03923}, 2024.

\bibitem{hao2023gnot}
Z.~Hao, Z.~Wang, H.~Su, C.~Ying, Y.~Dong, S.~Liu, Z.~Cheng, J.~Song, and J.~Zhu, ``{GNOT: A general neural operator transformer for operator learning},'' in \emph{International Conference on Machine Learning}.\hskip 1em plus 0.5em minus 0.4em\relax PMLR, 2023, pp. 12\,556--12\,569.

\bibitem{he2023mgno}
J.~He, X.~Liu, and J.~Xu, ``Mgno: Efficient parameterization of linear operators via multigrid,'' \emph{arXiv preprint arXiv:2310.19809}, 2023.

\bibitem{starr1969nonzero}
A.~W. Starr and Y.-C. Ho, ``Nonzero-sum differential games,'' \emph{Journal of optimization theory and applications}, vol.~3, no.~3, pp. 184--206, 1969.

\bibitem{bressan2010noncooperative}
A.~Bressan, ``Noncooperative differential games. a tutorial,'' \emph{Department of Mathematics, Penn State University}, vol.~81, 2010.

\bibitem{bui2022optimizeddp}
M.~Bui, G.~Giovanis, M.~Chen, and A.~Shriraman, ``Optimizeddp: An efficient, user-friendly library for optimal control and dynamic programming,'' \emph{arXiv preprint arXiv:2204.05520}, 2022.

\bibitem{leung2020infusing}
K.~Leung, E.~Schmerling, M.~Zhang, M.~Chen, J.~Talbot, J.~C. Gerdes, and M.~Pavone, ``On infusing reachability-based safety assurance within planning frameworks for human--robot vehicle interactions,'' \emph{The International Journal of Robotics Research}, vol.~39, no. 10-11, pp. 1326--1345, 2020.

\bibitem{fridovich2020confidence}
D.~Fridovich-Keil, A.~Bajcsy, J.~F. Fisac, S.~L. Herbert, S.~Wang, A.~D. Dragan, and C.~J. Tomlin, ``Confidence-aware motion prediction for real-time collision avoidance1,'' \emph{The International Journal of Robotics Research}, vol.~39, no. 2-3, pp. 250--265, 2020.

\bibitem{jacot2018neural}
A.~Jacot, F.~Gabriel, and C.~Hongler, ``Neural tangent kernel: Convergence and generalization in neural networks,'' \emph{Advances in neural information processing systems}, vol.~31, 2018.

\bibitem{lee2019wide}
J.~Lee, L.~Xiao, S.~Schoenholz, Y.~Bahri, R.~Novak, J.~Sohl-Dickstein, and J.~Pennington, ``Wide neural networks of any depth evolve as linear models under gradient descent,'' \emph{Advances in neural information processing systems}, vol.~32, 2019.

\bibitem{wang2022and}
S.~Wang, X.~Yu, and P.~Perdikaris, ``{When and why PINNs fail to train: A neural tangent kernel perspective},'' \emph{Journal of Computational Physics}, vol. 449, p. 110768, 2022.

\bibitem{mohamadi2023fast}
M.~A. Mohamadi, W.~Bae, and D.~J. Sutherland, ``A fast, well-founded approximation to the empirical neural tangent kernel,'' in \emph{International Conference on Machine Learning}.\hskip 1em plus 0.5em minus 0.4em\relax PMLR, 2023, pp. 25\,061--25\,081.

\bibitem{singarcs}
E.~Cristiani and P.~Martinon, ``Initialization of the shooting method via the hamilton-jacobi-bellman approach,'' \emph{Journal of Optimization Theory and Applications}, vol. 146, no.~2, pp. 321--346, 2010.

\bibitem{so2025solving}
O.~So, C.~Ge, and C.~Fan, ``Solving minimum-cost reach avoid using reinforcement learning,'' \emph{Advances in Neural Information Processing Systems}, vol.~37, pp. 30\,951--30\,984, 2025.

\end{thebibliography}
\bibliographystyle{IEEEtran}
\end{document}